\documentclass[11pt,a4paper]{article}
\usepackage[hyperref]{emnlp-ijcnlp-2019}
\usepackage{times}
\usepackage{latexsym}
\usepackage{enumerate}
\usepackage{graphicx}
\usepackage{subfigure}
\usepackage{amsmath}
\usepackage{amsfonts}
\usepackage{multirow}
\usepackage{bm}
\usepackage{array}
\usepackage{verbatim}
\usepackage{xcolor}
\usepackage{url}
\usepackage{color}
\usepackage{enumitem}
\usepackage{verbatim}
\aclfinalcopy 

\title{CAN: Constrained Attention Networks for Multi-Aspect Sentiment Analysis}

\author{Mengting Hu\textsuperscript{1}\thanks{\; This work was done when Mengting Hu was a research intern at IBM Research - China.} \quad Shiwan Zhao\textsuperscript{2}\thanks{\; Corresponding author.} \quad Li Zhang\textsuperscript{2} \quad Keke Cai\textsuperscript{2}\\
{\bf Zhong Su\textsuperscript{2} \quad Renhong Cheng\textsuperscript{1} \quad Xiaowei Shen\textsuperscript{2}} \\
\textsuperscript{1} Nankai University \quad \textsuperscript{2} IBM Research - China \\
mthu@mail.nankai.edu.cn, \{zhaosw, lizhang, caikeke, suzhong\}@cn.ibm.com \\ chengrh@nankai.edu.cn, xwshen@cn.ibm.com
}

\date{}

\begin{document}
\maketitle
\begin{abstract}
Aspect level sentiment classification is a fine-grained sentiment analysis task. To detect the sentiment towards a particular aspect in a sentence, previous studies have developed various attention-based methods for generating aspect-specific sentence representations. However, the attention may inherently introduce noise and downgrade the performance. In this paper, we propose constrained attention networks (CAN), a simple yet effective solution, to regularize the attention for multi-aspect sentiment analysis, which alleviates the drawback of the attention mechanism. Specifically, we introduce orthogonal regularization on multiple aspects and sparse regularization on each single aspect. 
Experimental results on two public datasets demonstrate the effectiveness of our approach. We further extend our approach to multi-task settings and outperform the state-of-the-art methods.
\end{abstract}

\section{Introduction}
\begin{figure}
\centering
\includegraphics[width=0.45\textwidth]{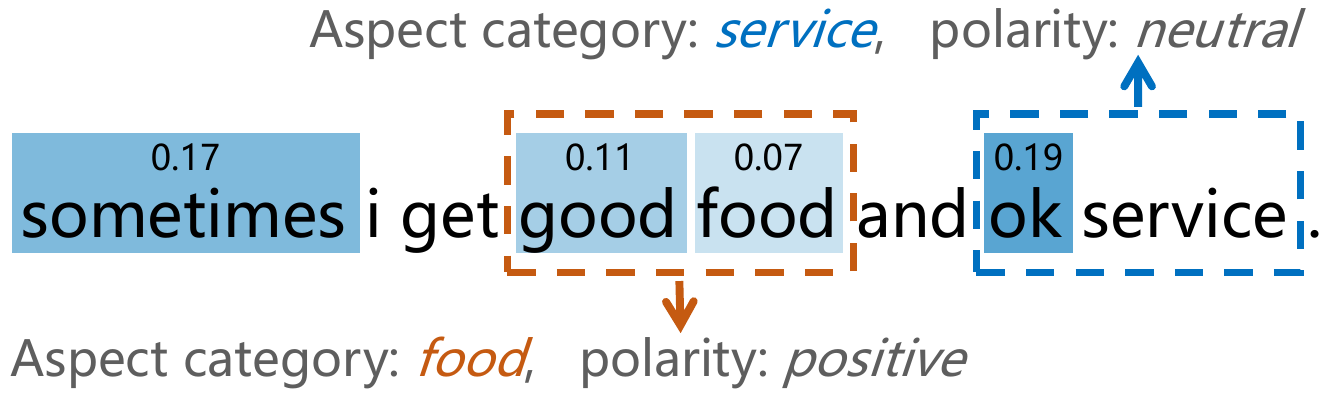}
\caption{Example of a non-overlapping sentence. The attention weights
of the aspect \emph{food} are from the model ATAE-LSTM \cite{Wang2016Attention}.} 
  \label{sentence} 
\end{figure}

Sentiment analysis \cite{Nasukawa2003Sentiment,liu2012sentiment}, an important task in natural language understanding, receives much attention in recent years. Aspect level sentiment classification (ALSC) is a fine-grained sentiment analysis task, which aims at detecting the sentiment towards a particular aspect in a sentence. ALSC is especially critical for multi-aspect sentences which contain multiple aspects. A multi-aspect sentence can be categorized as \emph{overlapping} or \emph{non-overlapping}. A sentence is annotated as non-overlapping if any two of its aspects have no overlap. Our study found that around $85\%$ of the multi-aspect sentences are non-overlapping in the two public datasets. Figure \ref{sentence} shows a simple example. The non-overlapping sentence contains two aspects. The aspect \emph{food} is on the left side of the aspect \emph{service}. Their distributions on words are {\bf orthogonal} to each other. Another observation is that only a few words relate to the opinion expression in each aspect. As shown in Figure \ref{sentence}, only the word \emph{``good''} is relevant to the aspect \emph{food} and \emph{``ok"} to \emph{service}. The distribution of the opinion expression of each aspect is {\bf sparse}. 

To detect the sentiment towards a particular aspect, previous studies \cite{Wang2016Attention,Ma2017Interactive,cheng2017aspect,ma2018targeted,huang2018aspect} have developed various attention-based methods for generating aspect-specific sentence representations. In these works, the attention may inherently introduce noise and downgrade the performance \cite{li2018acl} since the attention scatters across the whole sentence and is prone to attend on noisy words, or the opinion words from other aspects. Take Figure \ref{sentence} as an example, for the aspect \emph{food}, we visualize the attention weights from the model \cite{Wang2016Attention}. Much of the attention focuses on the noisy word \emph{``sometimes"}, and the opinion word \emph{``ok"} which is relevant to the aspect \emph{service} rather than \emph{food}.

To alleviate the above issue, we propose a model for multi-aspect sentiment analysis, which regularizes the attention by handling multiple aspects of a sentence simultaneously. Specifically, we introduce orthogonal regularization for attention weights among multiple non-overlapping aspects. The orthogonal regularization tends to make the attention weights of multiple aspects concentrate on different parts of the sentence with less overlap. We also introduce the sparse regularization, which tends to make the attention weights of each aspect concentrate only on a few words. We call our networks with such regularizations \emph{constrained attention networks} (CAN). 
There have been some works on introducing sparsity in attention weights in machine translation \cite{sparse-NMT} and orthogonal constraints in domain adaptation \cite{DSN_nips}. In this paper, we add both sparse and orthogonal regularizations in a unified form inspired by the work \cite{lin2017structured}.
The details will be introduced in Section \ref{sec:model}.

In addition to aspect level sentiment classification (ALSC), aspect category detection (ACD) is another task of aspect based sentiment analysis.  ACD~\cite{Zhou2015Representation,Schouten2018Supervised} aims to identify the aspect categories discussed in a given sentence from a predefined set of aspect categories (e.g., price, food, service). Take Figure \ref{sentence} as an example, aspect categories \emph{food} and \emph{service} are mentioned. We introduce ACD as an auxiliary task to assist the ALSC task, benefiting from the shared context of the two tasks. We also apply our attention constraints to the ACD task. By applying attention weight constraints to both ALSC and ACD tasks in an end-to-end network, we can further evaluate the effectiveness of CAN in multi-task settings.

In summary, the main contributions of our work are as follows:
\begin{itemize}
\item We propose CAN for multi-aspect sentiment analysis. Specifically, we introduce orthogonal and sparse regularizations to constrain the attention weight allocation, helping learn better aspect-specific sentence representations. 

\item We extend CAN to multi-task settings by introducing ACD as an auxiliary task, and applying CAN on  both ALSC and ACD tasks. 

\item Extensive experiments are conducted on public datasets. Results demonstrate the effectiveness of our approach for aspect level sentiment classification.  
\end{itemize}

\section{Related Work}
{\bf Aspect level sentiment analysis} is a fine-grained sentiment analysis task. Earlier methods are usually based on explicit features \cite{liu2010improving,Vo2015Target}. 
With the development of deep learning technologies, various neural attention mechanisms have been proposed to solve this fine-grained task~\cite{Wang2016Attention,ruder2016hierarchical,Ma2017Interactive,tay2017dyadic,cheng2017aspect,RA-emnlp-2017,Tay2018Learning,ma2018targeted,wang2018learning,wang2018clause}. To name a few,  \newcite{Wang2016Attention} propose an attention-based LSTM network for aspect level sentiment
classification. 
\newcite{Ma2017Interactive} use the interactive attention
networks to generate the representations for targets and contexts separately. 
\newcite{cheng2017aspect,ruder2016hierarchical} both propose hierarchical neural network models for aspect level sentiment classification.  
\newcite{wang2018learning} employ a segmentation
attention based LSTM model for aspect level sentiment classification. All these works can be categorized as single-aspect sentiment analysis, which deals with aspects in a sentence separately, without considering the relationship between aspects.

More recently, a few works have been proposed to take the relationship among multiple aspects into consideration. \newcite{multi-aspect1-naacl} make simultaneous classification of all aspects in a sentence using recurrent networks.  \newcite{multi-aspect3-emnlp} employ memory network to model the dependency of the target aspect with the other aspects in the sentence. \newcite{multi-aspect2-emnlp} design an aspect alignment loss to enhance the difference of the
attention weights towards the aspects which have
the same context and different sentiment polarities. In this paper, we introduce orthogonal regularization to constrain the attention weights of multiple non-overlapping aspects, as well as sparse regularization on each single aspect.

\begin{figure*}[ht]
\centering
	\includegraphics[width=1.0\textwidth]{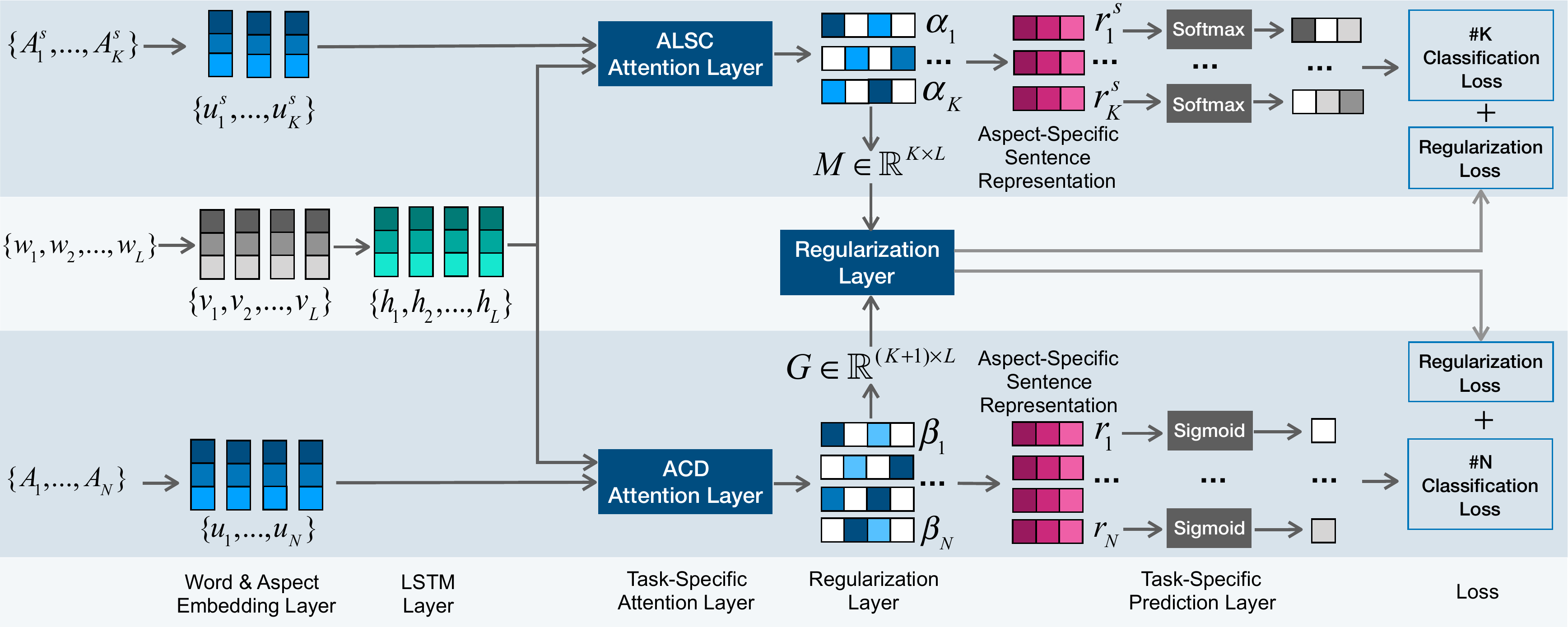}
    \caption{Network Architecture. The aspect categories are embedded as vectors. The model encodes the sentence using LSTM. Based on its hidden states, aspect-specific sentence representations for ALSC and ACD tasks are learned via constrained attention. Then aspect level sentiment prediction and aspect category detection are made. }
    \label{network}
\end{figure*}

{\bf Multi-task learning} \newcite{Caruana1997Multitask} solves multiple learning tasks at the same time, achieving improved performance by exploiting commonalities and differences across tasks. Multi-task learning has been used successfully in many  machine learning applications. \newcite{Huang2018Multitask} learn both main task and auxiliary task jointly with shared representations, achieving improved performance in question answering. \newcite{Toshniwal2017Multitask} use low-level auxiliary tasks
for encoder-decoder based speech recognition, which suggests that the addition of auxiliary tasks can help in either optimization or generalization. \newcite{yu2016learning} use two auxiliary tasks to help induce a sentence embedding that works well across domains for sentiment classification. In this paper, we adopt the multi-task learning approach by using ACD as the auxiliary task to help the ALSC task. 


\section{Model}
\label{sec:model}
We first formulate the problem. There are totally $N$ predefined aspect categories in the dataset, $A=\{A_1,...,A_N\}$. Given a sentence $S=\{w_1, w_2, ..., w_L\}$, which contains $K$ aspects $A^s=\{A_1^s,...,A_K^s\}, A^s\subseteq  A$, the multi-task learning is to simultaneously solve the ALSC and ACD tasks, namely, the ALSC task predicts the sentiment polarity of each aspect $A_k^s \in A^s$, and the auxiliary ACD task checks each aspect $A_n \in A$ to see whether the sentence $S$ mentions it. Note that we only focus on aspect-category instead of aspect-term \cite{xue2018acl} in this paper.

We propose CAN for multi-aspect sentiment analysis, supporting both ALSC and ACD tasks by a multi-task learning framework. The network architecture is shown in Figure \ref{network}. We will introduce all components sequentially from left to right.

\subsection{Embedding and LSTM Layers}
Traditional single-aspect sentiment analysis handles each aspect separately, one at a time. In such settings, a sentence $S$ with $K$ aspects will be copied to form $K$ instances. For example, a sentence $S$ contains two aspects: $A_1^s$ with polarity $p_1$ and $A_2^s$ with polarity $p_2$. Two instances, $\langle{S, A_1^s, p_1}\rangle$ and $\langle{S, A_2^s, p_2}\rangle$, will be constructed. Our multi-aspect sentiment analysis method handles multiple aspects together and takes the single instance $\langle{S, [A_1^s, A_2^s], [p_1, p_2]}\rangle$ as input. 

The input sentence $\{w_1, w_2, ..., w_L\}$ is first converted to a sequence of vectors $\{v_1,v_2,...,v_L\}$, and the $K$ aspects of the sentence are transformed to vectors $\{u_1^s,...,u_K^s\}$, which is a subset of $\{u_1,...,u_N\}$, the vectors of all aspect categories. The word embeddings of the sentence are then fed into an LSTM network \cite{Hochreiter1997Long}, which outputs hidden states $H=\{h_1,h_2,...,h_L\}$. The sizes of the embedding and the hidden state are both set to be $d$.

\subsection{Task-Specific Attention Layer}
The ALSC and ACD tasks share the hidden states from the LSTM layer, while compute their own attention weights separately. The attention weights are then used to compute aspect-specific sentence representations. 

{\bf ALSC Attention Layer}
The key idea of aspect level sentiment classification is to learn different attention weights for different aspects, so that different aspects can concentrate on different parts of the sentence. 
We follow the approach in the work \cite{Bahdanau2015iclr} to compute the attention. Particularly, given the sentence $S$ with $K$ aspects, $A^s=\{A_1^s,...,A_K^s\}$, for each aspect $A_k^s$, its attention weights are calculated by:
\begin{equation}
    \alpha_k = softmax({z^a}^\mathrm{T}tanh(W_{1}^{a}{H} + W_{2}^{a}(u_k^{s}\otimes{e_L}))) 
  \label{equation_absa_att}
\end{equation}
where $u_k^{s}$ is the embedding of the aspect $A_k^s$, $e_L\in\mathbb{R}^{L}$ is a
vector of $1$s, ${u_k^{s}}\otimes{e_L}$ is the operation repeatedly concatenating $u_k^{s}$ for $L$ times. $W_{1}^a\in\mathbb{R}^{{d}\times{d}}$, $W_{2}^a\in\mathbb{R}^{{d}\times{d}}$ and $z^a\in\mathbb{R}^{d}$ are the weight matrices.

{\bf ACD Attention Layer} We treat the ACD task as multi-label classification problem for the set of $N$ aspect categories. For each aspect $A_n\in A$, its attention weights are calculated by:
\begin{equation}
    \beta_n = softmax({z^b}^\mathrm{T}tanh(W_{1}^{b}{H} + W_{2}^{b}(u_n\otimes{e_L}))) 
  \label{equation_acd_att}
\end{equation}
where $u_n$ is the embedding of the aspect $A_n$. $W_{1}^b\in\mathbb{R}^{{d}\times{d}}$, $W_{2}^b\in\mathbb{R}^{{d}\times{d}}$ and $z^b\in\mathbb{R}^{d}$ are the weight matrices.

The ALSC and ACD tasks use the same attention mechanism, but they do not share parameters. The reason to use separated parameters is that, for the same aspect, the attention of ALSC concentrates more on opinion words, while ACD focuses more on aspect target terms (see the attention visualizations in Section~\ref{sec:att_vis}). 

\subsection{Regularization Layer}
We simultaneously handles multiple aspects by adding constraints to their attention weights. {\bf Note that this layer is only available in the training stage}, in which the ground-truth aspects are known for calculating the regularization loss, and then influence parameter updating in back propagation. While in the testing/inference stage, the true aspects are unknown and the regularization loss is not calculated so that this layer is omitted from the architecture.  

In this paper, we introduce two types of regularizations: the sparse regularization on each single aspect; the orthogonal regularization on multiple non-overlapping aspects. 

{\bf Sparse Regularization} For each aspect, the sparse regularization constrains the distribution of the attention weights ($\alpha_k$ or $\beta_n$) to concentrate on less words. For simplicity, we use $\alpha_k$ as an example, $\alpha_k=\{\alpha_{k1},  \alpha_{k2}, ..., \alpha_{kL}\}$. To make $\alpha_k$ sparse, the sparse regularization term is defined as: 
\begin{equation}
    R_s = \mid\sum\limits_{l=1}^L{\alpha_{kl}^{2}} - 1\mid
    \label{equation:sparse_reg}
\end{equation}

where $\sum\limits_{l=1}^L{\alpha_{kl}}=1$ and $\alpha_{kl}>0$. Since $\alpha_k$ is normalized as a probability distribution, $L_1$ norm is always equal to $1$ (the sum of the probabilities) and does not work as sparse regularization as usual. Minimizing Equation \ref{equation:sparse_reg} will force the sparsity of $\alpha_k$. It has the similar effect as minimizing the entropy of $\alpha_k$, which leads to placing more probabilities on less words.

{\bf Orthogonal Regularization} This regularization term forces orthogonality among attention weight vectors of multiple aspects, so that different aspects attend on different parts of the sentence with less overlap. Note that we only apply this regularization to non-overlapping multi-aspect sentences. Assume that the sentence $S$ contains $K$ non-overlapping aspects $\{A_1^s,...,A_K^s\}$ and their attention weight vectors are $\{\alpha_1,...,\alpha_K\}$. We pack them together as a two-dimensional attention matrix $M\in\mathbb{R}^{{K}\times{L}}$ to calculate the orthogonal regularization term.
\begin{equation}
	R_o = \parallel{M^{\mathrm{T}}}M - I\parallel_2
\end{equation}
where $I$ is an identity matrix. In the resulted matrix of ${M^{\mathrm{T}}M}$, each non-diagonal element is the dot product between two attention weight vectors, minimizing the non-diagonal elements will force orthogonality between corresponding attention weight vectors. The diagonal elements of ${M^{\mathrm{T}}M}$ are subtracted by $1$, which are the same as $R_s$ defined in Equation \ref{equation:sparse_reg}. As a whole, $R_o$ includes both sparse and orthogonal regularization terms. 

Note that in the ACD task, we do not pack all the $N$ attention vectors $\{\beta_1, ..., \beta_N\}$ as a matrix. The sentence $S$ contains $K$ aspects. For simplicity, let $\{\beta_1, ..., \beta_K\}$ be the attention vectors of the $K$ aspects mentioned, while $\{\beta_{K+1}, ..., \beta_N\}$ be the attention vectors of the $N-K$ aspects not mentioned. We compute the average of the $N-K$ attention vectors, denoted by $\beta_{avg}$. We then construct the attention matrix $G=\{\beta_{1}, ..., \beta_{K},\beta_{avg}\}$, $G\in\mathbb{R}^{{(K+1)}\times{L}}$. The reason why we calculate $\beta_{avg}$ is that if an aspect is not mentioned in the sentence, its attention weights often attend to meaningless stop words, such as \emph{``to''}, \emph{``the''}, \emph{``was''}, etc. We do not need to distinguish among the $N-K$ aspects not mentioned, therefore they can share stop words in the sentence by being averaged as a whole, which keeps the $K$ aspects mentioned away from such stop words. 

\subsection{Task-Specific Prediction Layer}
Given the attention weights of each aspect, we can generate aspect-specific sentence representation, and then make prediction for the ALSC and ACD tasks respectively.

{\bf ALSC Prediction} The weighted hidden state is combined with the last hidden state to generate the final aspect-specific sentence representation.
\begin{equation} 
    r_k^s=tanh(W_1^r{\bar{h}_k}+W_2^rh_L)
\end{equation}
where $W_1^r\in\mathbb{R}^{{d}\times{d}}$ and $W_2^r\in\mathbb{R}^{{d}\times{d}}$. $\bar{h}_k =\sum\limits_{l=1}^L{\alpha_{kl}h_l}$ is the weighted hidden state for aspect $k$. $r_k^s$ is then used to make sentiment polarity prediction.
\begin{equation} 
    \hat{y_k}=softmax(W_p^{a}r_k^s+b_p^{a})
\end{equation}
where $W_p^a\in\mathbb{R}^{{d}\times{c}}$ and $b_p^a\in\mathbb{R}^{{c}}$ are the parameters of the projection layer, and $c$ is the number of classes. 

For the sentence $S$ with $K$ aspects mentioned, we make $K$ predictions simultaneously. That is why we call our approach multi-aspect sentiment analysis.  

{\bf ACD Prediction} We directly use the weighted hidden state as the sentence representation for ACD prediction.
\begin{equation} 
r_n =\bar{h}_n =\sum\limits_{l=1}^L{\beta_{nl}h_l} 
\end{equation} 
We do not combine with the last hidden state $h_L$ since the aspect may not be mentioned by the sentence. We make $N$ predictions for all predefined aspect categories. 
\begin{equation} 
    \hat{y_n}=sigmoid(W_p^{b}r_n+b_p^{b})
\end{equation}
where $W_p^{b}\in\mathbb{R}^{{d}\times{1}}$ and $b_p^{b}$ is a scalar. 

\subsection{Loss}
For the task ALSC, the loss function for the $K$ aspects of the sentence $S$ is defined by:
\begin{equation} \nonumber
	L_a = -\sum\limits_{k=1}^{K}\sum\limits_{c}y_{kc}\log\hat{y_{kc}}
    \label{equation_absa_loss}
\end{equation}
where $c$ is the number of classes. For the task ACD, as each prediction is binary classification problem, the loss function for the $N$ aspects of the sentence $S$ is defined by:
\begin{equation}\nonumber
	L_b = -\sum\limits_{n=1}^{N}[y_n\log{\hat{y_n}} + (1-y_n)\log(1-\hat{y_n})]
    \label{equation_acd_loss}
\end{equation}

We jointly train our model for the two tasks. The
parameters in our model are then trained by minimizing the combined loss function:
\begin{equation}
	L = L_a + \frac{1}{N}L_b + \lambda{R} 
    \label{equation_loss_together}
\end{equation}
where $R$ is the regularization term mentioned previously, which can be $R_s$ or $R_o$. $\lambda$ is the hyperparameter used for tuning the impact from regularization
loss to the overall loss. To avoid $L_b$ overwhelming the overall loss, we divide it by the number of aspect categories.

\begin{table}[t!]
\begin{center}
\setlength{\tabcolsep}{1mm}{
\begin{tabular} {|c|c|ccc|c|}
\hline
    \multirow{2}{*}{Dataset} & \multirow{2}{*}{\#Single} & \multicolumn{3}{c|}{\#Multi} & \multirow{2}{*}{\#Total} \\
    \cline{3-5}
      &&  \emph{OL} & \emph{NOL} & \emph{Total} & \\
	\hline
		Rest14\_Train & 2053 & 67 & 415 & 482 & 2535\\
        Rest14\_Val & 412 & 19 & 75 & 94 & 506 \\
		Rest14\_Test & 611 & 27 & 162 & 189 & 800 \\
        \hline
        Rest15\_Train & 622 & 47 & 262 & 309 & 931 \\
        Rest15\_Val & 137 & 13 & 39 & 52 & 189 \\
        Rest15\_Test & 390 & 30 & 162 & 192 & 582 \\
	\hline
\end{tabular}}
\end{center}
\caption{\label{table-dataset} The numbers of single- and multi-aspect sentences. \emph{OL} and \emph{NOL} denote the overlapping and non-overlapping multi-aspect sentences, respectively.}
\end{table}

\section{Experiments}
\subsection{Datasets}
We conduct experiments on two public datasets from SemEval 2014 task 4 \cite{Pontiki2014SemEval} and SemEval 2015 task 12 \cite{semeval-2015} (denoted by Rest14 and Rest15 respectively). These two datasets consist of restaurant customer reviews with annotations identifying the mentioned aspects and the sentiment polarity of each aspect. To apply orthogonal regularization, we manually annotate the multi-aspect sentences with overlapping or non-overlapping\footnote{We will release the annotated dataset later.}. We randomly split the original training set into training, validation sets in the ratio 5:1, where the validation set is used to select the best model. We count the sentences of single-aspect and multi-aspect separately. Detailed statistics are summarized in Table \ref{table-dataset}. Particularly, $85.23\%$ and $83.73\%$ of the multi-aspect sentences are non-overlapping in Rest14 and Rest15, respectively. 

\begin{table*}[t!]
\begin{center}
\setlength{\tabcolsep}{2.0mm}{
\begin{tabular} {|c|cc|cc|cc|cc|}
\hline
	\multirow{2}{*}{Model} & \multicolumn{4}{c|}{Rest14} & \multicolumn{4}{c|}{Rest15} \\ 
    \cline{2-9}
    & \multicolumn{2}{c|}{3-way} & \multicolumn{2}{c|}{Binary} & \multicolumn{2}{c|}{3-way} & \multicolumn{2}{c|}{Binary} \\ 
    \cline{2-9}
    & Acc & F1 & Acc & F1 & Acc & F1 & Acc & F1 \\
	\hline
    LSTM      & 80.92 & 68.30 & 85.83 & 80.88 & 71.24 & 49.40 & 71.97 & 69.97 \\ 
    AT-LSTM   & 81.24 & 69.19 & 87.25 & 82.20 & 73.37 & 51.74 & 76.79 & 74.61 \\
    ATAE-LSTM & 82.18 & 69.18 & 88.08 & 83.03 & 74.56 & 51.40 & 79.79 & 78.69 \\
    GCAE      & 82.08 & 70.20 & 87.72 & 83.84 & 76.69 & 53.00 & 79.66 & 77.96 \\
    \hline
    AT-CAN-$R_s$   & 82.28 & 70.94 & 88.43 & 84.07 & 75.62 & 53.56 & 78.36 & 76.69 \\
    AT-CAN-$R_o$   & 82.81 & 71.32 & {\bf 89.37} & {\bf 85.66} & 76.92 & {\bf 55.67} & 79.92 & 78.77 \\
    ATAE-CAN-$R_s$ & 81.97 & 72.19 &  88.90 & 84.29 & 77.28 & 52.45 & 81.49 & 80.61 \\
    ATAE-CAN-$R_o$ & {\bf 83.33} & {\bf 73.23} & 89.02 & 84.76 & {\bf 78.58} & 54.72 & {\bf 81.75} & {\bf 80.91}  \\
	\hline
\end{tabular}}
\end{center}
\caption{\label{table-st} Results of the ALSC task in single-task settings in terms of accuracy ($\%$) and Macro-F1 ($\%$).}
\end{table*}

\subsection{Comparison Methods}
Since we focus on aspect-category sentiment analysis, many works \cite{Ma2017Interactive,li2018acl,multi-aspect2-emnlp} which focus on aspect-term sentiment analysis are excluded.
\begin{itemize}[leftmargin=*]
\item {\bf LSTM}: We implement the vanilla LSTM to model the sentence and use the average of all hidden states as the sentence representation. In this model, aspect information is not used.

\item {\bf AT-LSTM} \cite{Wang2016Attention}: It adopts the attention mechanism in LSTM to generate a weighted representation of a sentence. The aspect embedding is used to compute the attention weights as in Equation \ref{equation_absa_att}. We do not concatenate the aspect embedding to the hidden state as in the work \cite{Wang2016Attention} and gain small performance improvement. We use this modified version in all experiments. 

\item {\bf ATAE-LSTM} \cite{Wang2016Attention}: This method is an extension of AT-LSTM. In this model, the aspect embedding is concatenated to each word embedding of the sentence as the input to the LSTM layer. 


\item {\bf GCAE} \cite{xue2018acl}: This state-of-the-art method is based on the convolutional
neural network with gating mechanisms, which is for both aspect-category and aspect-term sentiment analysis. We compare with its aspect-category sentiment analysis task.

\end{itemize}

\subsection{Our Methods}
To verify the performance gain of introducing constraints on attention weights, we first create several variants of our model for single-task settings.

\begin{itemize}[leftmargin=*]

\item {\bf AT-CAN-\boldmath{$R_s$}}: Add sparse regularization $R_s$ to AT-LSTM to constrain the attention weights of each single aspect. 

\item {\bf AT-CAN-\boldmath{$R_o$}}: Add orthogonal regularization $R_o$ to AT-CAN-$R_s$ to constrain the attention weights of multiple non-overlapping aspects. 

\item {\bf ATAE-CAN-\boldmath{$R_s$}}: Add $R_s$ to ATAE-LSTM.

\item {\bf ATAE-CAN-\boldmath{$R_o$}}: Add $R_o$ to ATAE-CAN-$R_s$.
\end{itemize}

We then extend attention constraints to multi-task settings, creating variants by different options: 1) no constraints, 2) adding regularizations only to the ALSC task, 3) adding regularizations to both tasks.
\begin{itemize}[leftmargin=*]

\item {\bf M-AT-LSTM}: This is the basic multi-task model without regularizations.

\item {\bf M-CAN-\boldmath{$R_s$}}: Add $R_s$ to the ALSC task in M-AT-LSTM.

\item {\bf M-CAN-\boldmath{$R_o$}}: Add $R_o$ to the ALSC task in M-CAN-$R_s$.

\item {\bf M-CAN-2\boldmath{$R_s$}}: Add $R_s$ to both tasks in M-AT-LSTM. 

\item {\bf M-CAN-2\boldmath{$R_o$}}: Add $R_o$ to both tasks in M-CAN-2$R_s$.
\end{itemize}

\begin{table*}[t!]
\begin{center}
\setlength{\tabcolsep}{2.0mm}{
\begin{tabular} {|c|cc|cc|cc|cc|}
\hline
	\multirow{2}{*}{Model} & \multicolumn{4}{c|}{Rest14} & \multicolumn{4}{c|}{Rest15} \\ 
    \cline{2-9}
    & \multicolumn{2}{c|}{3-way} & \multicolumn{2}{c|}{Binary} & \multicolumn{2}{c|}{3-way} & \multicolumn{2}{c|}{Binary} \\
    \cline{2-9}
    & Acc & F1 & Acc & F1 & Acc & F1 & Acc & F1 \\
	\hline
    M-AT-LSTM    & 82.60 & 71.44 & 88.55 & 83.76 & 76.33 & 51.64 & 79.53 & 78.31 \\
	M-CAN-$R_s$  & 83.65 & 73.97 & 89.26 & 85.43 & 75.74 & 52.43 & 79.66 & 78.46 \\
    M-CAN-$R_o$  & 83.12 & 72.29 & 89.61 & 85.18 & 77.04 & 52.69 & 79.40 & 77.88 \\
    M-CAN-2$R_s$ & 83.23 & 72.81 & 89.37 & 85.42 & {\bf 78.22} & {\bf 55.80} & 80.44 & 80.01 \\
    M-CAN-2$R_o$ & {\bf 84.28} & {\bf 74.45} & {\bf 89.96} & {\bf 86.16} & 77.51 & 52.78 & {\bf 82.14} & {\bf 81.58} \\
\hline
\end{tabular}}
\end{center}
\caption{\label{table-mt} Results of the ALSC task in multi-task settings in terms of accuracy ($\%$) and Macro-F1 ($\%$).}
\end{table*}

\begin{table*}[t!]
\begin{center}
\setlength{\tabcolsep}{2.0mm}{
\begin{tabular} {|c|ccc|ccc|}
\hline
	\multirow{2}{*}{Model} & \multicolumn{3}{c|}{Rest14} & \multicolumn{3}{c|}{Rest15} \\ 
	\cline{2-7}
	& Precision & Recall & F1 & Precision & Recall & F1 \\ 
	\hline
    M-AT-LSTM     & 0.8626 & 0.8553 & 0.8589 & 0.6691 & 0.4748 & 0.5555\\
	M-CAN-2$R_s$  & 0.8698 & 0.8595 & 0.8645 & 0.6244 & {\bf 0.5019} & 0.5565\\
    M-CAN-2$R_o$  & {\bf 0.8907} & {\bf 0.8627} & {\bf 0.8765} & {\bf 0.7127} & 0.4865 & {\bf 0.5782}\\
\hline
\end{tabular}}
\end{center}
\caption{\label{table-acd} Results of the ACD task. Rest14 has $5$ aspect categories while Rest15 has $13$ ones.}
\end{table*}

\subsection{Implementation Details}
We set $\lambda=0.1$ with the help of the validation set. All models are optimized by the Adagrad optimizer \cite{duchi2011adaptive} with learning rate 0.01. Batch size is 25. We apply a dropout of $p=0.7$ after the embedding and LSTM layers. All words in the sentences are initialized with 300 dimension Glove Embeddings \cite{Pennington2014Glove}. The aspect embedding matrix and parameters are initialized by sampling from a uniform distribution $U(-\varepsilon,\varepsilon)$, $\varepsilon=0.01$. $d$ is set as 300. The models are trained for 100 epochs, during which the model with the best performance on the validation set is saved. We also apply early stopping in training, which means that the training will stop if the performance on validation set does not improve in 10 epochs. 


\subsection{Results}
Table \ref{table-st} and \ref{table-mt} show our experimental results on the two public datasets for single-task and multi-task settings respectively. In both tables, ``3-way'' stands for 3-class classification (positive, neutral, and negative), and ``Binary'' for binary classification (positive and negative). The best scores are marked in bold.

{\bf Single-task Settings} Table \ref{table-st} shows our experimental results of ALSC in single-task settings. Firstly, we observe that by introducing attention regularizations (either $R_s$ or $R_o$), most of our proposed methods outperform their counterparts. Particularly, AT-CAN-$R_s$ and AT-CAN-$R_o$ outperform AT-LSTM in all results; ATAE-CAN-$R_s$ and ATAE-CAN-$R_o$ also outperform ATAE-LSTM in $15$ of $16$ results. For example, in the Rest15 dataset, ATAE-CAN-$R_o$ outperforms ATAE-LSTM by up to $5.39\%$ of accuracy and $6.46\%$ of the F1 score in the 3-way classification. Secondly, regularization $R_o$ achieves better performance improvement than $R_s$ in all results. This is because $R_o$ includes both orthogonal and sparse regularizations for non-overlapping multi-aspect sentences. Thirdly, our approaches, especially ATAE-CAN-$R_o$, outperform the state-of-the-art baseline model GCAE. Finally, the LSTM method outputs the worst results in all cases, because it can not distinguish different aspects.

{\bf Multi-task Settings} Table \ref{table-mt} shows experimental results of ALSC in multi-task settings. We first observe that the overall results in multi-task settings outperform the ones in single-task settings, which demonstrates the effectiveness of multi-task learning by introducing the auxiliary ACD task to help the ALSC task. Second, in almost all cases, applying attention regularizations to both tasks gains more performance improvement than only to the ALSC task, which shows that our attention regularization approach can be extended to different tasks which involving aspect level attention weights, and works well in multi-task settings. For example, for the Binary classification in the Rest15 dataset, M-AT-LASTM outperforms AT-LSTM by $3.57\%$ of accuracy and $4.96\%$ of the F1 score, and M-CAN-2$R_o$ further outperforms M-AT-LSTM by $3.28\%$ of accuracy and $4.0\%$ of the F1 score. 


Table \ref{table-acd} shows the results of the ACD task in multi-task settings. Our proposed regularization terms can also improve the performance of ACD. Regularization $R_o$ achieves the best performance in almost all metrics. 

\subsection{Attention Visualizations}
\label{sec:att_vis}

\begin{figure}[ht]
\centering
\subfigure[AT-LSTM]{
\includegraphics[width=0.45\textwidth]{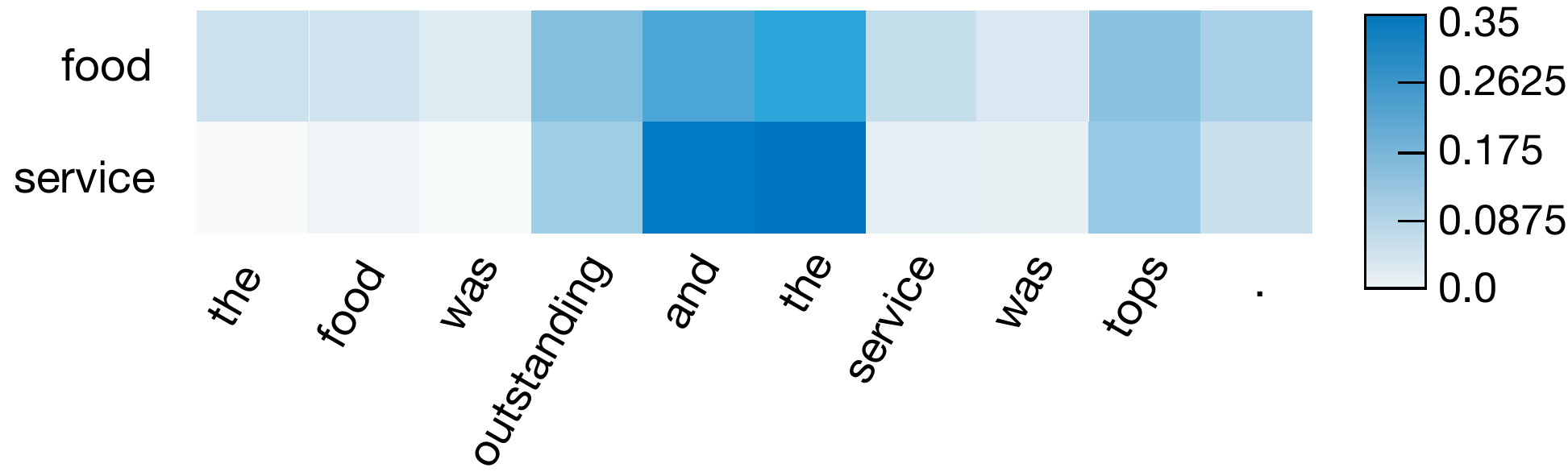}}
\subfigure[M-AT-LSTM]{
\includegraphics[width=0.45\textwidth]{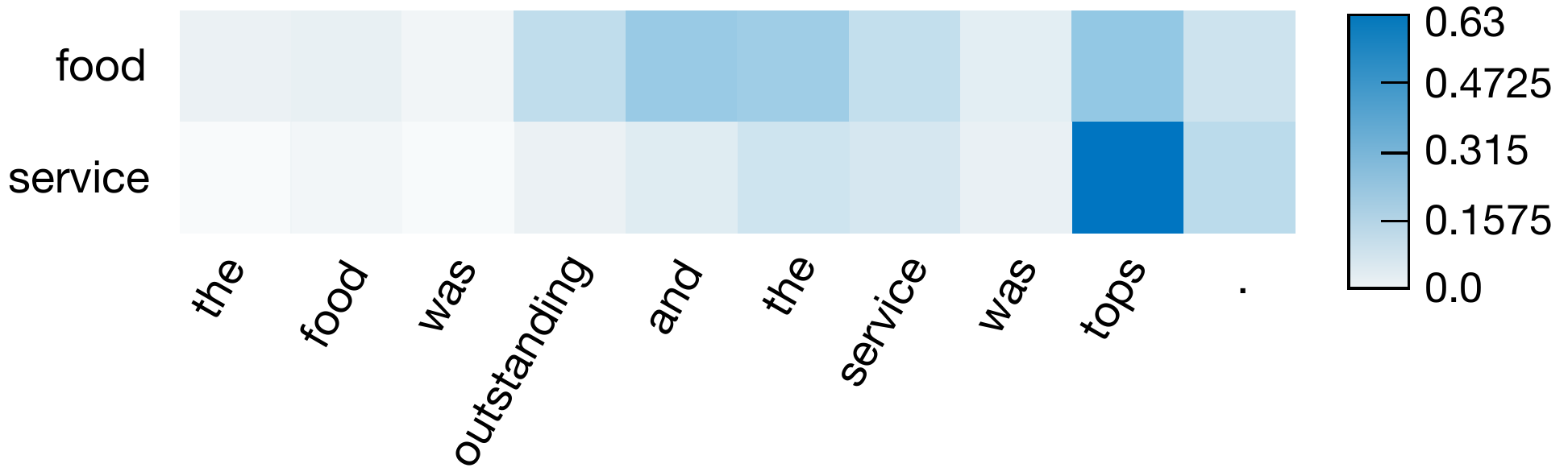}}
\subfigure[M-CAN-2$R_o$]{
\includegraphics[width=0.45\textwidth]{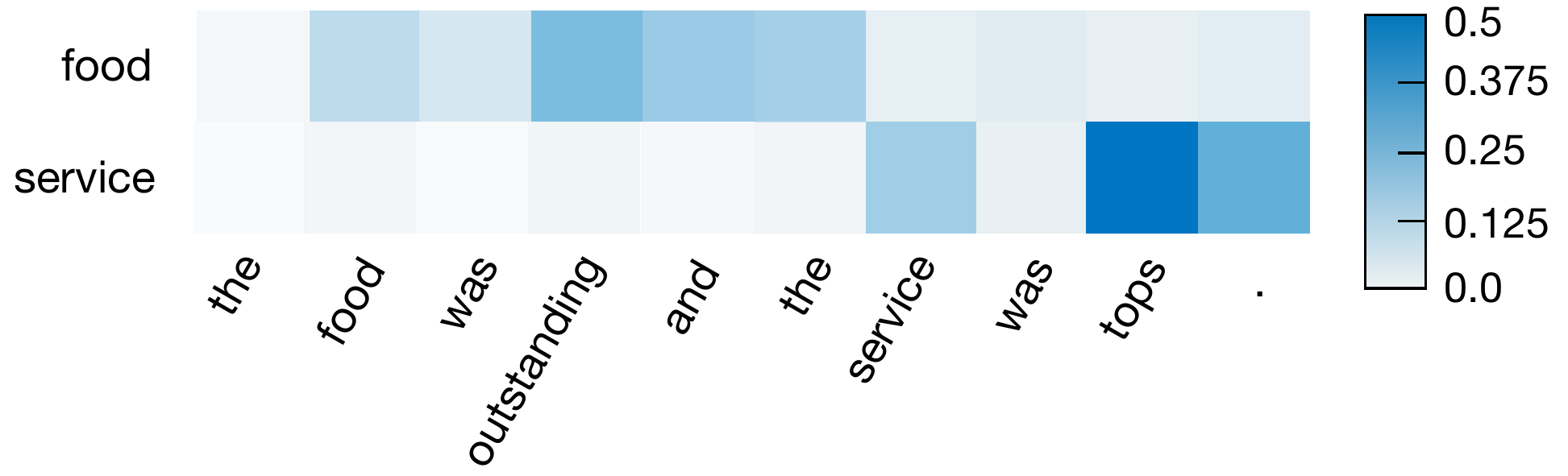}}
\caption{Visualization of attention weights of different aspects in the ALSC task. Three different models are compared.}
\label{compare-att}
\end{figure}

\begin{figure}[ht]
\centering
\includegraphics[width=0.45\textwidth]{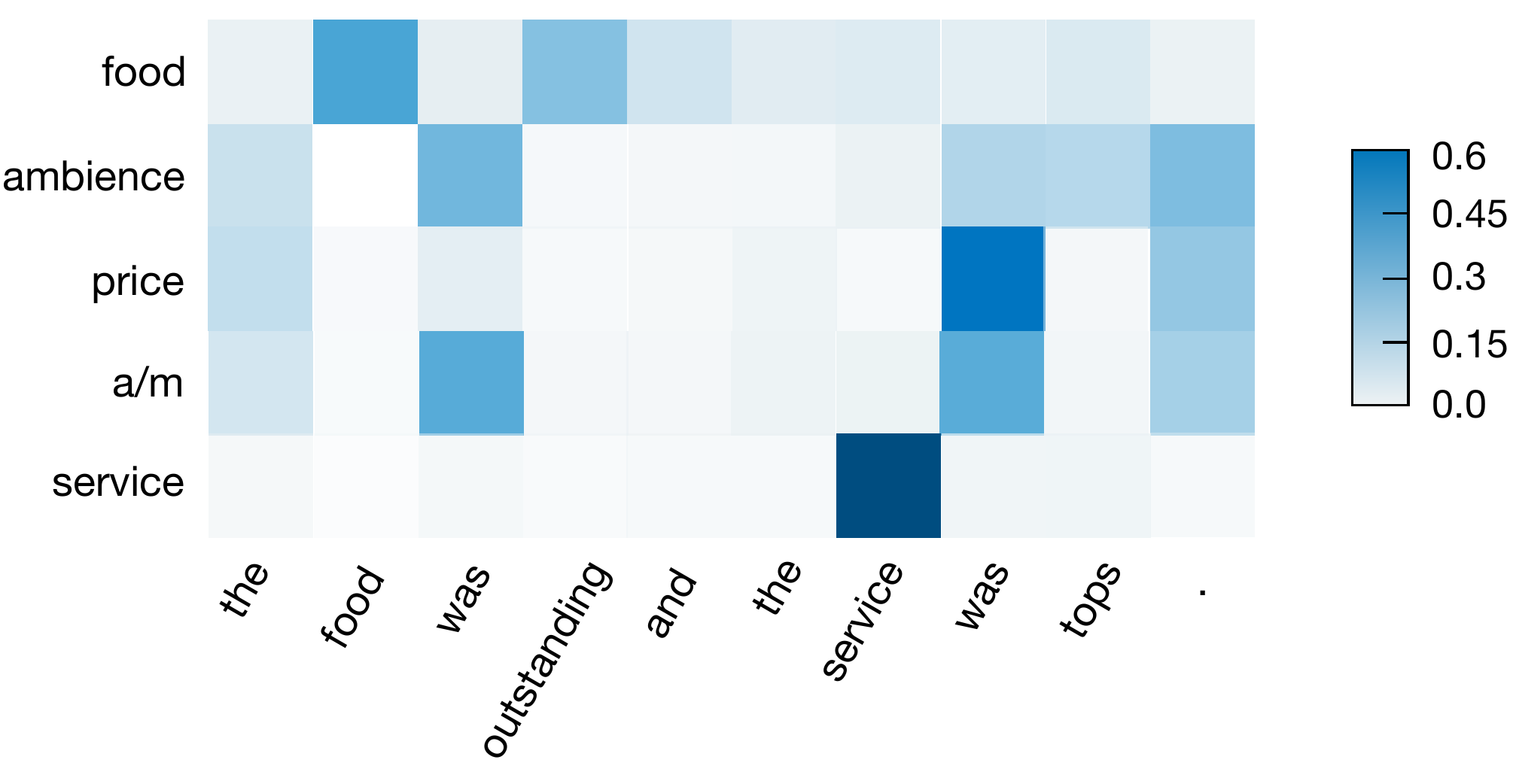}
\caption{Visualization of attention weights of different aspects in the ACD task from M-CAN-2$R_o$. The a/m is short for anecdotes/miscellaneous.} 
  \label{ACD-att} 
\end{figure}

\begin{figure}
\centering
\includegraphics[width=0.45\textwidth]{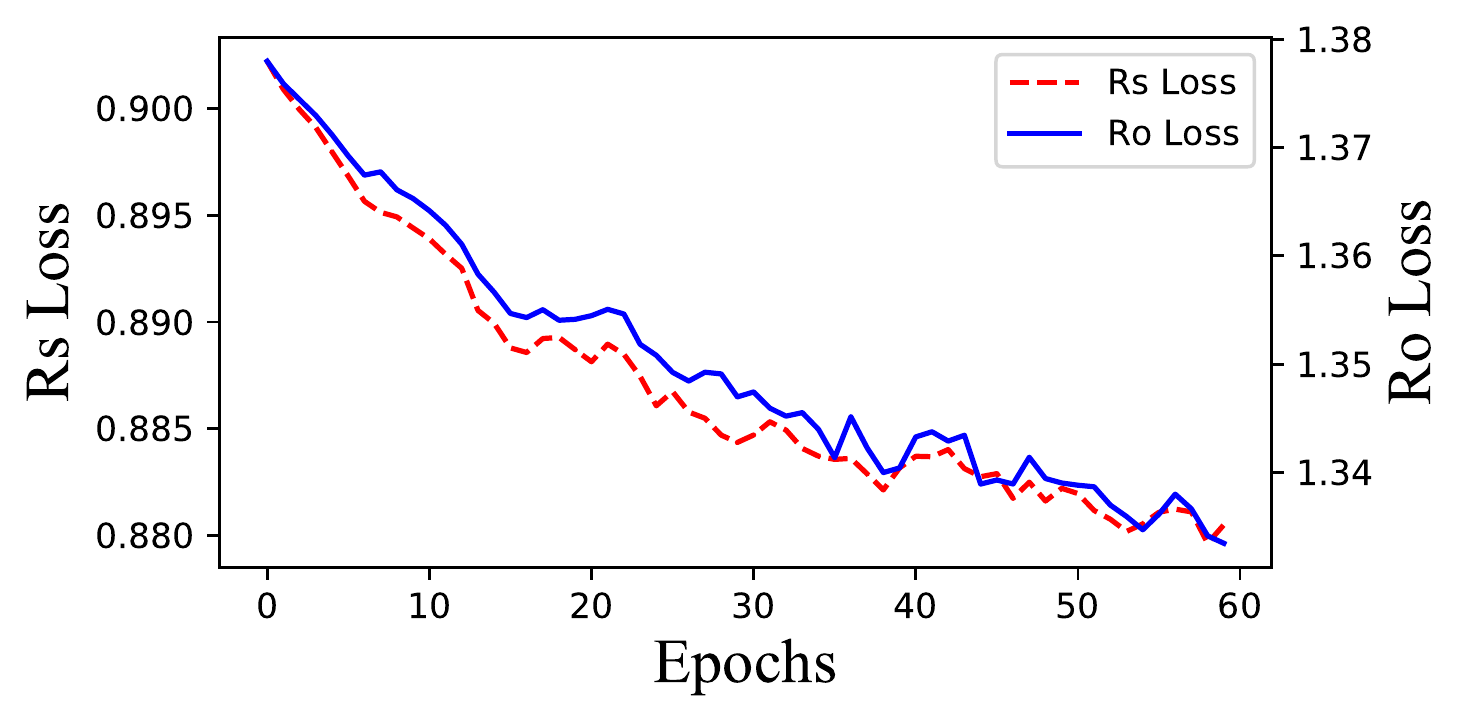}
\caption{The regularization loss curves of $R_s$ and $R_o$ during the training of AT-CAN-$R_o$.} 
  \label{figure:reg-loss} 
\end{figure}

Figure \ref{compare-att} depicts the attention weights from AT-LSTM, M-AT-LSTM and M-CAN-2$R_o$ methods, which are used to predict the sentiment polarity in the ALSC task. The subfigure (a), (b) and (c) show the attention weights of the same sentence, for the aspect \emph{food} and \emph{service} respectively. We observe that the attention weights of each word associated with each aspect are quite different for different methods. For AT-LSTM method in subfigure (a), the attention weights of aspect \emph{food} and \emph{service} are both high in words \emph{``outstanding''}, \emph{``and''}, and \emph{``the''}, but actually, the word \emph{``outstanding''} is used to describe the aspect \emph{food} rather than \emph{service}. The same situation occurs with the word \emph{``tops''}, which should associate with
\emph{service} rather than \emph{food}. The attention mechanism alone is not good enough to locate aspect-specific opinion words and generate aspect-specific sentence representations in the ALSC task.
\newline\indent
As shown in subfigure (b), the issue is mitigated in M-AT-LSTM. Multi-task learning can learn better hidden states of the sentence, and better aspect embeddings. However, it is still not good enough. For instance, the attention weights of the word \emph{``tops''} are both high for the two aspects, and the weights are overlapped in the middle part of the sentence. 
\newline\indent
As shown in subfigure (c), M-CAN+2$R_o$ generates the best attention weights. The attention weights of the aspect \emph{food} are almost orthogonal to the weights of \emph{service}. The aspect \emph{food} concentrates on the first part of the sentence while \emph{service} on the second part. Meanwhile, the key opinion words \emph{``outstanding''} and \emph{``tops''} get highest attention weights in the corresponding aspects. 
\newline\indent
We also visualize the attention for the auxiliary task ACD. Figure \ref{ACD-att} depicts the attention weights from the method M-CAN-2$R_o$. There are five predefined aspect categories (\emph{food, ambience, price, anecdotes/miscellaneous, service}) in the dataset, two of which are mentioned in the sentence. In the ACD task, we need to calculate the attention weights for all the five aspect categories, and then generate aspect-specific sentence representations to determine whether the sentence contains each aspect. As shown in Figure \ref{ACD-att}, attention weights for aspects \emph{food} and \emph{service} are pretty good. The aspect \emph{food} concentrates on words \emph{``food''} and \emph{``outstanding''}, and the aspect \emph{service} focuses on the word \emph{``service''}. It is interesting that for aspects which are not mentioned in the sentence, their attention weights often attend to meaningless stop words, such as \emph{``the''}, \emph{``was''}, etc. We do not distinguish these aspects and just treat them as a whole.
\newline\indent
We plot the regularization loss curves in Figure \ref{figure:reg-loss}, which shows that both $R_s$ and $R_o$ decrease during the training of AT-CAN-$R_o$.

\begin{figure*}
\centering
	\includegraphics[width=0.99\textwidth]{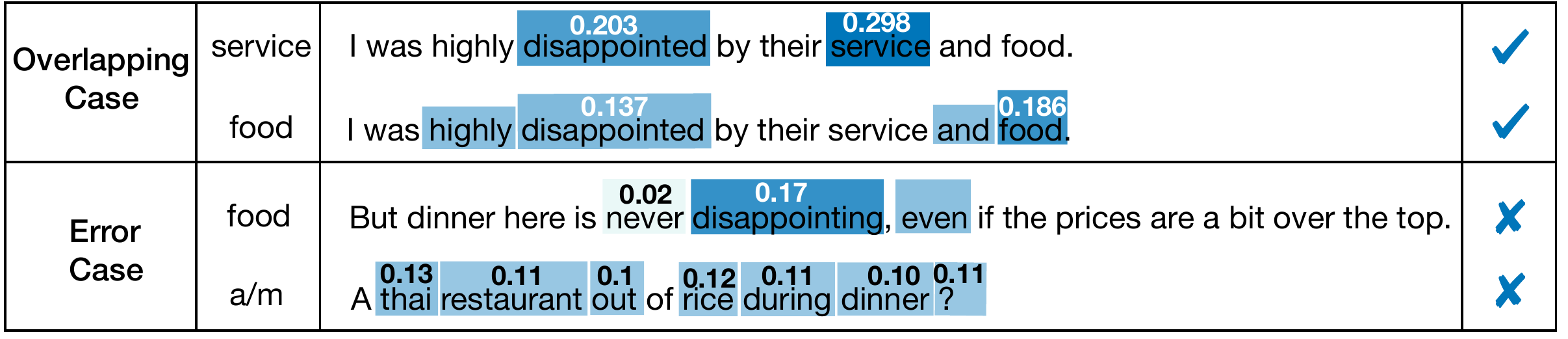}
    \caption{Examples of overlapping case and error case. The a/m is short for anecdotes/miscellaneous.}
    \label{case}
\end{figure*}

\subsection{Case Studies}
{\bf Overlapping Case} We only add sparse regularization to overlapping sentences in which multiple aspects share the same opinion snippet. As shown in Figure \ref{case}, the sentence contains two aspects \emph{food} and \emph{service}, both described by the opinion snippet \emph{``highly disappointed''}. Our method can locate the aspect terms and shared opinion words for both aspects, and then classify the sentiment correctly. 

{\bf Error Case} With the help of attention visualization, we can conduct error analysis of our model conveniently. As shown in Figure \ref{case}, for the first sentence in error case, the aspect \emph{food} attends on the right word \emph{``disappointing''}, but fails to include the negation word \emph{``never''}. This may be caused by the inaccurate sentence representation or aspect embedding. We can not rebuild the connection between the aspect and the word by our regularizations. The second sentence is negative but classified as neutral. The attention weights distribute evenly since the sentence does not contain any explicit opinion words. Since there is no tendency which words to concentrate, our model can not adjust the attention weights and help on such cases.

\section{Conclusions}
We propose constrained attention networks for multi-aspect sentiment analysis, which handles multiple aspects of a sentence simultaneously. Specifically, we introduce orthogonal and sparse regularizations on attention weights. Furthermore, we introduce an auxiliary task ACD for promoting the ALSC task, and apply CAN on both tasks. Experimental results demonstrate that our approach outperforms the state-of-the-art methods. 

\section{Acknowledgement}
This work is supported by National Science and Technology Major Project, China (Grant No. 2018YFB0204304).

\bibliography{emnlp-ijcnlp-2019}
\bibliographystyle{acl_natbib}

\end{document}